**Downscaling human mobility data based on demographic, socioeconomic, and commuting characteristics using interpretable machine learning methods**

**Authors:**
**Yuqin Jiang\*,** Department of Geography and Environment, University of Hawaii at Manoa, Honolulu, Hawaii, USA
**Andrey A. Popov,** Department of Information and Computer Science, University of Hawaii at Manoa, Honolulu, Hawaii, USA
**Tianle Duan,** School of Construction Management Technology, Purdue University, West Lafayette, Indiana, USA
**Qingchun Li**, School of Construction Management Technology, Purdue University, West Lafayette, Indiana, USA

\*Corresponding author: yuqinj@hawaii.edu

**Abstract**
Understanding urban human mobility patterns at various spatial levels is essential for social science. This study presents a machine learning framework to downscale origin-destination (OD) taxi trips flows in New York City from a larger spatial unit to a smaller spatial unit. First, correlations between OD trips and demographic, socioeconomic, and commuting characteristics are developed using four models: Linear Regression (LR), Random Forest (RF), Support Vector Machine (SVM), and Neural Networks (NN). Second, a perturbation-based sensitivity analysis is applied to interpret variable importance for nonlinear models. The results show that the linear regression model failed to capture the complex variable interactions. While NN performs best with the training and testing datasets, SVM shows the best generalization ability in downscaling performance. The methodology presented in this study provides both analytical advancement and practical applications to improve transportation services and urban development.

# 1. Introduction

Understanding human mobility patterns is crucial for various aspects of urban study. These patterns inform local business strategies, transportation planning, and urban development. However, in modern metropolitan areas, human mobility patterns have become increasingly complex, presenting challenges for analysis and prediction[1–4]. Existing research has sought to unravel these complex mobility patterns by examining the demographic and socioeconomic factors associated with travel behaviors [5–7]. Traditionally, this has been approached through two primary methods: active and passive data collection. Active data collection methods, such as surveys and interviews, provide detailed insights into individual travel preferences and behaviors, but are often limited in scale and representativeness [8–10]. Passive methods, on the other hand, leverage data from sources such as cell phone or vehicle-based travel data, offering broader coverage but frequently resulting in aggregated datasets with coarse spatial resolution due to privacy concerns and access limitations [2, 11, 12].

The analytical approaches used to interpret travel behaviors also face limitations. Traditional statistical methods, such as regression models, are useful for identifying linear correlations between variables and are straightforward to interpret. However, they struggle to capture more complicated non-linear relationships in mobility patterns. More advanced methods, such as Random Forest, Support Vector Machine and Neural Network models can handle non-linear correlations, but are harder to interpret for practical applications [13–16].

This study aims to address these methodological challenges by leveraging interpretable machine learning and GeoAI techniques. Specifically, our research has two primary objectives:
1. To develop non-linear methods for downscaling spatial units based on demographic and socioeconomic variables, which can improve the granularity of mobility analysis.
2. To identify key demographic, socioeconomic and commuting characteristics that have the most significant impacts on travel patterns, providing a more comprehensive understanding of influential factors affecting travel patterns.

To achieve these goals, we utilize New York City taxi datasets as our case study. We first develop non-linear models to establish correlations between demographic, socioeconomic, and commuting characteristics and the number of trips between origin-destination pairs. The non-linear models are usually considered as a black box, where the influence of each independent variable is not directly provided. To tackle this problem, we apply a perturbation-based sensitivity method to interpret these models and derive the pseudo-coefficients, which is similar to the linear coefficients derived from the linear regression model. By doing so, we can identify the most influential demographic and socioeconomic features that affect the mobility patterns from these non-linear models. By combining advanced interpretable machine learning and GeoAI techniques with rich urban mobility datasets, this study aims to contribute to a more comprehensive understanding of human mobility patterns in complex urban environments.

This paper is organized as follows. Section 2 reviews related literature on the impact of demographic and socioeconomic characteristics on daily travel patterns, as well as existing studies on spatial data downscaling. Section 3 introduces the study area and the datasets used in this study. Section 4 provides a detailed description on methodology. Section 5 presents the results and discusses these findings. Lastly, section 6 concludes this paper.

## 2. Literature review
*2.1. The impact of demographic and socioeconomic characteristics on daily travel patterns*
Household income significantly influences daily travel patterns by shaping individuals' choices of transportation modes, trip frequency, travel distance, and destinations. Lower-income households often rely on public transit, such as buses, subways, and trains, due to limited access to private vehicles. For example, a study based on 2017 data found that only 51% of low-income households in New York State owned private vehicles, compared to 82% of higher-income households that owned at least one vehicle [17]. Additionally, low-income households are more sensitive to fluctuations in gas prices. Rising fuel costs may cause them to switch from driving to public transit and reduce trip frequency [6, 18, 19]. While public transit offers a more affordable alternative, it often requires longer travel times [20–22]. In contrast, higher-income households are more likely to own private vehicles, enabling longer travel distances, more frequent trips, and access to a wider range of destinations [5, 23]. Income levels also influence engagement in active travel modes, such as walking and biking. While lower-income individuals may walk more frequently due to financial constraints, higher-income groups tend to rely on private vehicles. However, a few studies suggested that in densely populated urban areas, such as New York City, higher-income individuals may live closer to workplaces and choose active transportation modes to reduce commuting time [3, 24, 25].

Racial and ethnic identity is another important factor shaping the daily mobility patterns in U.S. cities. This is particularly evident in the transportation mode choices. Studies have found that Hispanics and African Americans rely on public transit more than other racial groups, as they comprise 62% of bus riders in U.S. urban areas [10, 26, 27]. In addition, different racial and ethnic groups present different patterns in space usage and access to points-of-interests. Studies have shown that spatial segregation exists in daily mobility patterns as residents of predominantly white neighborhoods would avoid areas where African Americans or Hispanics are the predominant residents [28–30]. At the same time, studies also indicate that different racial and ethnic groups have inequity in accessibility to green spaces [31–33], healthy food [34–37], and healthcare facilities [8, 9, 38, 39].

Commuting time and the transportation mode choice are also important factors shaping daily urban mobility patterns [40–43]. According to U.S. Census estimates, the national average one-way commuting time is 26.8 minutes in. However, commute durations tend to be longer in large cities, with New York City leading at 40.1 minutes one-way [44]. This extended duration is largely attributed to heavy traffic congestion and urban sprawl. At the same time, a substantial number of NYC commuters choose active transportation modes, such as biking and walking, or rely on public transit systems, including subways and buses [3, 45]. The interaction between commuting time and mode choice generates cascading effects on broad urban mobility patterns. For example, longer commutes by car contribute to severe traffic congestion, which further increases travel times. Meanwhile, public transit may be inaccessible for individuals living in suburban areas far from their workplaces [4, 46–48]. With the rise of remote and hybrid work styles, commuting patterns have also shifted since the COVID-19 pandemic. Studies have found that commuting frequency has decreased as many employees have the flexibility to work from home [49–51]. However, public transit ridership has significantly declined in cities, as more people prefer private vehicles or active transportation modes to minimize the risk of infection [52–55].

Besides income, race, and commuting patterns, other demographic and socioeconomic factors, such as age, gender, and car ownership, also influence daily mobility patterns [56–60]. These factors influence mobility patterns mainly in two different aspects. Firstly, individuals across different groups exhibit various destination interests, reflecting distinct interactions with the built urban environment. Secondly, these demographic and socioeconomic characteristics are closely associated with transportation mode choices, which further impacts travel frequency, travel time, and route choices [12, 61–63].

2.2. Spatial downscaling

Downscaling refers to the process of transforming data from a coarse spatial resolution to a finer resolution, enabling more detailed spatial analysis. Downscaling methods aim to estimate or predict values at finer spatial scales, revealing more detailed spatial patterns that are neglected at the larger spatial scale [64, 65]. Frequently, due to limitations in data collection, datasets are aggregated or averaged to relatively large spatial units, within which local variations are neglected. For example, demographic data from a national census typically report at a higher administrative level, such as counties. However, finer-level population estimates, such as at the block-level, are often needed for community-level planning [15, 66]. Similarly, environmental features derived from satellite, such as air temperature or soil moisture, are commonly available at a coarse raster resolution, but finer-resolution estimates are important for applications like small-scale agriculture and micro-climate modeling [13, 14, 67].

Many downscaling methods have been developed for raster datasets, as their regular grid-based spatial structure offers several analytical and computational advantages. Raster data provide uniformly sized and shaped spatial units, which can be easily implemented in algorithms [68]. Common downscaling methods include regression-based algorithms, support vector machines, random forest, and neural networks [13, 14, 16, 69]. Researchers have successfully downscaled many environmental and climate related raster datasets, such as temperature, precipitation, and soil moisture [13, 14, 67, 70]. Downscaled estimations play crucial roles in these fields as fine-resolution environmental datasets are essential for local-scale assessment, planning, and decision-making [68].

Compared to raster downscaling, downscaling vector datasets present greater challenges due to the complex geometric relationship and topological structure. Vector data are composed of polygons varying in shapes and sizes, making it difficult to apply uniform transformation rules [71–73]. To address these issues, vector-based downscaling methods often rely on correlations with multiple variables. These methods employ statistical, machine learning, or artificial intelligence methods to predict fine-scale values based on patterns learned from coarse spatial units. Several recent studies have explored vector-based downscaling. For example, Python et al. [74] applied Bayesian based statistical models to downscale COVID-19 case counts in China, considering factors including urban structure, disease spreading possibility, and population size. Kummu et al. [75] derived a high-resolution GDP database by employing machine learning algorithms that accounted for variables including urbanization level, travel time to city, and other economic measurements. In contrast to these single-location-based disaggregation studies, Tang et al. [76] expanded the downscaling method to origin-destination flows. Their study utilized gravity models to downscale spatial flows that incorporated distance decay effect, population, and economic factors at both the origin and the destination of the flows.

Building on these studies, our study also addresses the challenge of downscaling origin-destination flows. We incorporate a comprehensive set of variables at both the origin and the destination, including demographic, socioeconomic, and commuting-related factors. Our approach involves training statistical and machine learning models on the relationships observed at a coarse spatial scale. These trained models are then used to predict OD flows at a finer spatial scale, using the same set of independent variables available at the finer scale. Based on existing literature, we apply four statistical and machine learning methods: ordinary least squares (OLS) regression, random forest, support vector machine, and neural network. By validating the results generated by these models, our study offers insights into the relative strengths and limitations of each approach in downscaling OD flows.

### 3. Data and Study area

3.1. Study area

New York City (NYC) is the most populous city in the United States, home to more than 8.8 million residents within just over 300 square miles. The city is well-known for its high population density, averaging around 29,000 people per square mile (U.S. Census Bureau, n.d.). This population density, combined with a diverse and dynamic population, makes NYC a unique environment where efficient and flexible transportation is essential for daily life. Although the city has one of the most extensive public transit networks in the world, a substantial portion of NYC residents relies on taxis for convenience and comfort [2, 3].

NYC is composed of five boroughs, each of which is also a county in the state: Brooklyn (Kings County), Bronx (Bronx County), Manhattan (New York County), Queens (Queens County), and Staten Island (Richmond County). Manhattan is the economic and cultural core of the city, characterized by extremely high population density, iconic skyline, and famous landmarks, attracting millions of tourists each year [77]. Brooklyn and Queens are primarily residential areas with diverse ethical and cultural neighborhoods. The Bronx is located in the Northern part of the city. Staten Island is the most suburban of the five boroughs, with a relatively lower population density and limited public transit access compared to the rest of the city.

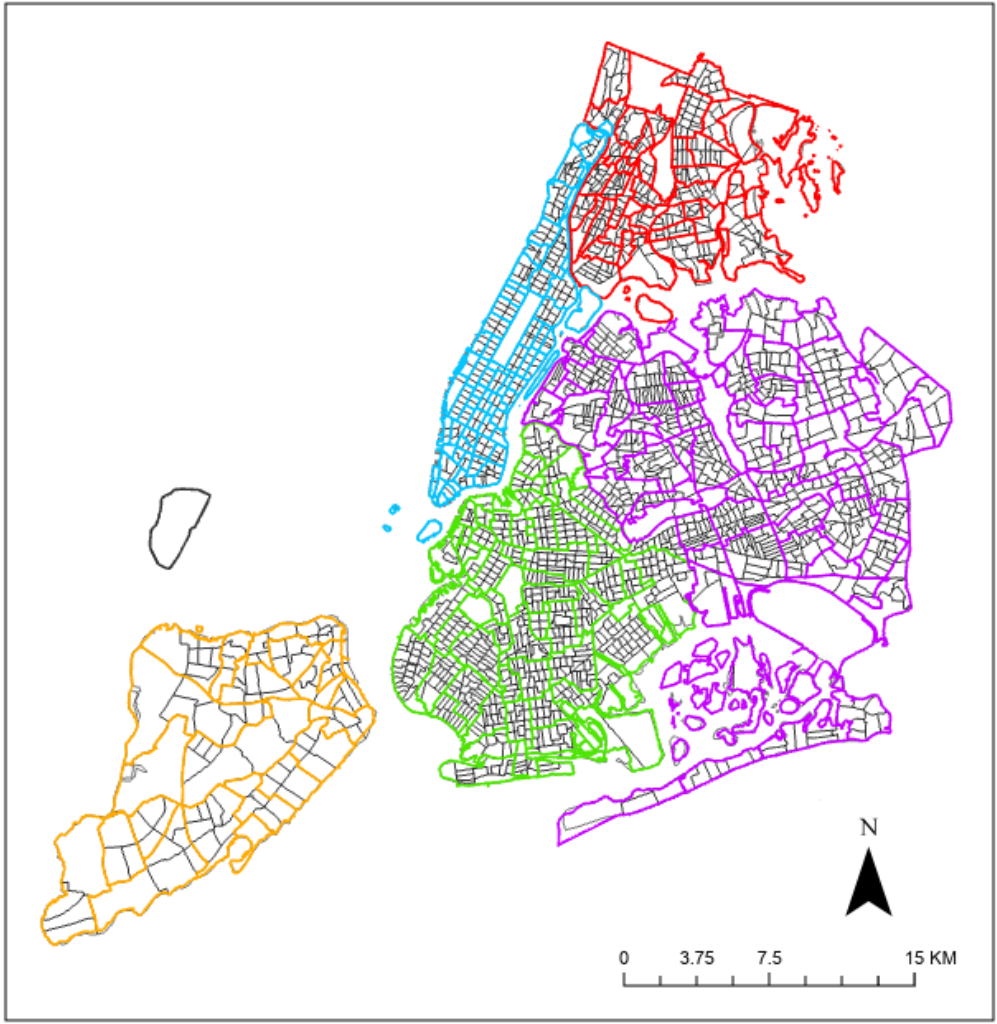

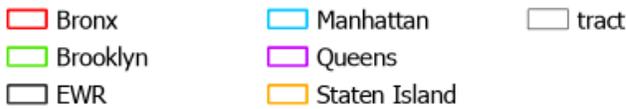

Figure 1. NYC taxi zones and tracts in the 5 boroughs and Newark Airport (EWR), NJ. The colored boundaries show the taxi zones in the 5 boroughs and the thin black lines indicate the boundaries for tracts in NYC.

3.2. Taxi flows

This study utilizes the publicly available New York City taxi trip records provided by the New York City Taxi & Limousine Commission [78]. The dataset contains detailed information on each trip, including pick-up and drop-off locations, date and time, passenger county, trip duration, trip distance, payment type, fare amount, surcharge, MTA tax, tolls, and total fare.

Notably, prior to 2015, the NYC TLC dataset included precise geographic coordinates for both pick-up and drop-off locations, allowing for detailed spatial analysis. To protect passenger privacy, the NYC TLC replaced the coordinates with 263 predefined taxi zones in the post-2015 datasets. In this study, we use the 2011 dataset, which retains the original geographic coordinates, enabling point-in-polygon aggregation to various spatial units. We first apply a spatial filter to exclude trips where either the origin or destination falls outside the NYC boundary. After this filtering step, we retain 135,582,947 valid trip records.

Using the spatial join operation, we aggregate individual trips into origin-destination (OD) flows between taxi zones, as well as between census tracts. Among the 263 taxi zones, 262 zones cover the five boroughs of NYC, while the remaining zone corresponds to Newark airport in New Jersey. Figure 1 shows the taxi zones and census tracts in NYC. These taxi zones closely align with the neighborhood boundaries, as shown in Figure 1. Each taxi zone contains one or more census tracts, making the taxi zone a relatively larger spatial unit and the tract a finer spatial unit. Since the boundaries of taxi zones and tracts are spatially aligned, we can harmonize the taxi trip data with demographic, socioeconomic, and commuting-related characteristics derived from the American Community Survey database. This spatial hierarchy provides the foundation for our downscaling models. We train the models using aggregated OD flows at the taxi-zone level and apply it to predict flows at the finer tract level.

3.3. Demographic and socioeconomic characteristics

The demographic, socioeconomic, and commuting-related data used in this study are obtained from the 2007–2011 five-year estimates of the American Community Survey (ACS). We downloaded these datasets through the National Historical Geographic Information System (NHGIS), a research platform maintained by the Minnesota Population Center at the University of Minnesota. NHGIS consolidates and provides access to standardized Census and ACS databases [79].

The selection of variables follows established literature on urban mobility and transportation modeling [3, 52, 80–83]. We include variables across several thematic categories: total population, total number of commuters, gender, age, race and ethnicity, commuting methods, commuting time, education attainment, economic measurements, housing condition, car ownership, and movements. A full list of the selected variables by each category is provided in Table 1.

Table 1. Selected demographic, socioeconomic and commuting characteristics.

| Category | variables |
|---|---|
| Population | Total population |
| | Total number of commuters |
| Age | Percentage of population that is 18 year-old or younger |
| | Percentage of population that is 60 year-old or older |
| Gender | Percentage of population that is male |
| Race and Ethnicity | Percentage of population that is white |
| | Percentage of population that is African American or black |
| | Percentage of population that is Asian |
| | Percentage of population that is Hispanic/Latino |
| Education | Percentage of population has a college degree or above |
| Economic | Percentage of population whose income is below poverty level in the past 12 months |
| | Percentage of population who was unemployment |
| | Median household income in the past 12 months |
| Housing | Percentage of housing units that is empty |
| | Percentage of housing units that is renter occupied |
| Stability | Percentage of population that lived in the same housing unit in the last 12 months |
| | Percentage of foreign-born population |
| Mobility | Percentage of households without a private vehicle |
| Commuting methods | Percentage of commuters who drives alone to work |
| | Percentage of commuters who carpools to work |
| | Percentage of commuters who uses bicycle as their main commuting method |
| | Percentage of commuters who uses motorcycle as their main commuting method |
| | Percentage of commuters who takes public transit to work |
| | Percentage of commuters who walks to work |
| | Percentage of commuters who takes taxi to work |
| | Percentage of commuters who work-from-home (no commuting) |
| Commuting time | Percentage of commuters whose commuting time is less than 30 minutes |
| | Percentage of commuters whose commuting time is between 30 and 60 minutes |
| | Percentage of commuters whose commuting time is between 60 and 90 minutes |
| | Percentage of commuters whose commuting time is more than 90 minutes |

All variables are initially processed at the tract level. To support model training and harmonize with the taxi dataset, we also aggregate these tract-level variables to the taxi zone level. Because each taxi zone contains one or more tracts and their boundaries are spatially aligned, this aggregation allows us to integrate ACS-derived features into both the training phase at the taxi one level and the downscaling phase at the tract level.

## 4. Methodology

4.1. Data processing

Demographic, socioeconomic and commuting related characteristics used in this study are derived from the ACS database. Notably, ACS data are tied to respondents' residential addresses, meaning that spatial units without permanent residents lack corresponding demographic, socioeconomic and commuting variables. In NYC, there are 42 tracts and 11 taxi zones with no permanent residents, including areas such as airports, parks, and natural reserves. However, many of those non-residential areas experience substantial population flows, including some high-traffic zones, such as Central Park, John F. Kennedy (JFK) and LaGuardia (LGA) airports. Although these areas do not have permanent residents, their population flows are highly influenced by individuals living in other parts of the city. Therefore, removing these areas or assigning zeros would fail to capture their roles in the urban movements and may distort the downscaling modeling. To address this issue, we apply a *mean imputation* method to fill in the missing demographic and socioeconomic variables. This method assigns the average value from all residential units to the non-residential units [84, 85]. For example, the "total population" of a non-residential tract is the mean of total population of all the residential tracts.

Next, the data are transformed to facilitate the application of various nonlinear methods in a consistent manner. Specifically, we normalize each variable using Z-score normalization using the following equation:

$$Z = \frac{X - \mu}{\sigma} \qquad \text{Eq. 1.}$$

In this way, the mean of each variable is zero and the standard deviation is one. Since the demographic, socioeconomic, and commuting-related variables vary in units and scales, direct inclusion without normalization can bias the modeling training process, particularly in models sensitive to feature magnitudes. Z-score normalization makes all variables unitless and puts them on the same scale. The normalization is applied to both the independent variables (X) and the dependent variable (Y). Following normalization, the dataset is split into training and validation sets, where 80% of the data are used to train the models and 20% of the data are reserved for evaluating model performance.

4.2. Training

In the model training process, our objective is to learn a mathematical function that can accurately predict the number of taxi flows (Y) based on a set of input variables (X). More specifically, we want to find a function, denoted as $g$, such that $Y = g(X) + \varepsilon$, where the goal is to find a $g$ such that the square of the error term $\varepsilon$ is minimized over all variables and all data points.

This prediction function $g$ produces predicted taxi flow amounts with the following equation: $Y_{predict} = g(X)$.

To evaluate how well a model performs, we calculate the Mean Squared Error (MSE), which is the average of the squared differences between predicted and actual values as:

$$MSE = \frac{1}{N}\sum_{i=1}^{N}\sum_{j=1}^{d}[g(Y_i^{test}) - Y_i^{predict}]_j^2, \qquad \text{Eq. 2.}$$

The training process aims to adjust the function $g$ in a way that minimizes this error across all data points, and the functions $g$ range over some set of functions that we define. Each machine learning model explored in this study (Random Forest, Support Vector Machine, and Neural Network) searches for this optimal function in its own way, based on the structure and flexibility allowed by the model.

A hyperparameter tuning process is conducted to ensure that each model generalizes well and does not overfit the training data. Hyperparameters are model-specific settings that influence how the prediction function $g$ is constructed but are not directly present in the formulas that define $g$. The goal is to find a combination of hyperparameters that minimizes prediction error but not overfit on the training data. In order to find the best hyperparameters, we perform a grid search over a wide range of hyperparameters for each of the non-linear methods and choose the combination that minimizes the error on the testing dataset. For instance, if one hyper-parameter is $a$ that is taken to be in the set {1, 2, 3} and another hyper-parameter is $b$ in the set {½, ⅓, ¼} then the grid search evaluates all nine possible combinations and selects the one that produces the lowest error on the testing dataset. This process helps identify the model settings that best balances between fitting the training data and performing well on the unseen testing data.

4.3. Models
We use 4 models in this study: multiple linear regression (MLR), random forest (RF), support vector machine (SVM), and neural network (NN). Among these, MLR is a linear model, which directly outputs the linear coefficients indicating how each independent variable can impact the dependent variable. On the other hand, RF, SVM, and NN are non-linear models. For these three non-linear models, we conduct a hyperparameter tuning to optimize the model performances. Then, we apply the optimal model to downscale the spatial. Lastly, we conduct a perturbation-based sensitivity analysis to identify the independent variables that most significantly influence the dependent variable.

4.3.1. Multiple Linear Regression
In order to establish a baseline for model performance, we begin with a linear regression model. The goal for this model is to predict the number of taxi trips (Y) using a linear combination of the input variables (X). The model takes the form $Y = b + AX + \varepsilon$, where $X$ is the matrix for independent variables. $A$ is a vector of coefficients, representing the importance of each variable. $b$ is the intercept term, and $\varepsilon$ is the error term. This formulation defines the prediction function as $g(X) = b + AX$, meaning that the relationship between inputs and output is assumed to be linear. The model parameters $A$ and $b$ are chosen by minimizing the sum of squared errors between the predicted and observed values. Although the linear regression provides an interpretable model, it assumes that the relationship between each predictor and the response is strictly linear, which limits its effectiveness in capturing more complex patterns in the data.

4.3.2. Random Forest

The first non-linear method we employ in this study is Random Forests (RF), an ensemble learning method that builds multiple decision trees and aggregates their predictions to produce a more robust output. Specifically, each tree in the forest is trained on a random subset of the training data, and the final prediction model is obtained by averaging the outputs of all trees. Mathematically, the function $g$ that is used by the random forest algorithm is defined as follows:

$$g(X) = \frac{1}{D}\sum_{d=1}^{D} t_d(X) \qquad \text{Eq. 3,}$$

where $D$ is the number of decision trees constructed and each $t_d$ is a separate decision tree. As each of the decision trees is only trained on a subset of the data, each tree has large variance in its predictions. By averaging the outputs across all trees, the RF can reduce variance and improves generalization performance.

The hyperparameters for RF used in this work are the **maximum number of features** (m), the **number of decision trees** (D), and **the random initialization**. The maximum number of features controls the complexity of the decision tree by determining the amount of branches at each split. It varies between two choices at each split: the square root of the number of variables and the base-2 logarithm of the number of variables. More features considered at each split allow the model to fit the data more closely, but may lead to overfitting problems, while fewer features can decrease the tree size and may avoid overfitting. The number of trees determines the size of the forest, which varies in the set of {10, 50, 100, 500, 1000}. The random initialization helps the algorithm consider different paths of learning that require random choices.

4.3.3. Support Vector Regression Model

The second non-linear model that is used in this work is Support Vector Regression (SVR). SVR is a variant of Support Vector Machine (SVM) designed for regression tasks. It aims to find a function that best approximates continuous target values while maintaining a margin of tolerance. The function $g$ that is used by the SVR algorithm is defined as follows,

$$g(X) = \sum_{i=1}^{N} a_i R(X, X_i) + b \qquad \text{Eq. 4,}$$

where $N$ are the number of training data points with each $X_i$ being a single training datum, the parameters $a_i$ and $b$ are chosen to minimize the error and satisfy the constraints of the support vector regression formulation, and, for this work, $R$ is the kernel function. In this work, we use the radial basis function (RBF) as the kernel function. The strength of SVR is that it is explicitly dependent on the training data, making it harder for the model to learn spurious relations that might exist in the data but not exist in the real world.

The hyperparameters that are tuned for this work are the **regularization coefficient** $C$, and the **maximum margin of tolerance** $e$. The hyperparameter $C$ is allowed to vary in the set {0.1, 1,

10} and the hyperparameter $e$ is allowed to vary in the set {0.01, 0.1, 1}. The regularization coefficient, $C$, for the support vector machine is a constant that modifies the cost function $g$ in a way to minimize the chance of overfitting. In general, a smaller C value means that the SVM does not fit the training dataset very well and allows more misclassifications. A larger C value means that the SVM fits the training data better, but may cause overfitting. Therefore, selecting an appropriate value is critical to achieving a good balance between fitting the training dataset well but avoiding overfitting. A delicate balance somewhere in the middle is required for the algorithm to generalize to the testing set without either being only useful on the training data, or being useless on all data. The maximum margin of tolerance, $e$, controls the effect of outliers on the resulting output. A small margin of tolerance means that outlier data points have little impact on the output, meaning that part of the sum of $g$ are only influenced by nearby data. A large margin of tolerance simply means that more data points are considered. In general, a smaller margin of tolerance can lead to more overfitting while a smaller margin of tolerance can lead to not generalizing well to the testing data.

### 4.3.4. Neural Network

The last non-linear model used in this study is Feedforward Neural Network (FNN). The hyperparameters of FNN not only define the architecture of the model but also influence the performance of it. In this study, the key hyperparameters include the **width**, **depth,** and **random initialization.** The width is the number of variables (neurons) in each hidden layer, which is varied in the set of {50, 100, 150}. The depth is the number of hidden layers in the network, which varies between the set of {2, 4}. The random initialization can impact convergence and performance and varies among five different sets of randomly initialized weights and biases. While some studies consider activation function as a hyperparameter in NN, in this study, it is fixed to **Rectified Linear Unit (ReLU)** due to its effectiveness and simplicity. For the training, the Adam optimization method is used, with the number of epochs is taken to be 2000, with a batch size of 5000, beta1 of 0.9 and beta2 of 0.95, and a triangular learning rate scheduler that varied the learning rate from a peak of $10^{-2}$ to a low of $10^{-6}$. A conceptual illustration of an FNN is shown in Figure 2. In this example, the network takes three input variables, processes them through two hidden layers, with five neurons at each hidden layer, and outputs a single prediction value.

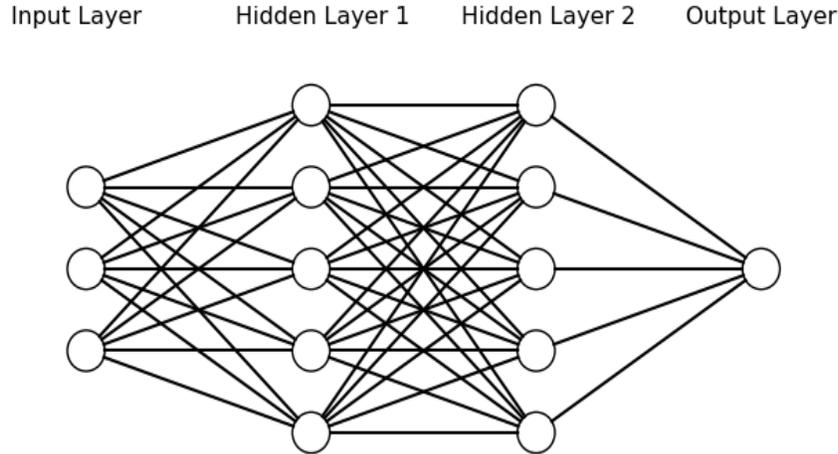
Figure 2. Illustration of an FNN architecture.

4.4. Downscaling to a smaller geographical unit

After developing the optimal non-linear models (RF, SVM, and FNN) at the taxi zone level, we apply each of them to the census tract level to predict taxi trip numbers at a finer spatial resolution. Each taxi trip involves an origin (O) and a destination (D), meaning that the prediction depends on demographic, socioeconomic, and commuting variables from both spatial units. We extract the 30 variables from each tract and combine them based on OD tract pairs. Similar to the model training process, for each independent variable (X-tract) and the dependent variable (Y-tract), we apply the z-score normalization to ensure consistency with the scale using Eq. 1.

Independent variables (X-tract) are then fed into the non-linear models, whose outputs are the predicted taxi trip number for each tract pair. The model predicts the number of trips between each pair of tracts based on the relationships it learned at the taxi zone level. To evaluate the performance of the downscaling, we calculate the MSE between the predicted trip numbers and the actual observed trip numbers using Eq. 2.

By incorporating both origin and destination variables, the model captures the bi-directional difference in demographic, socioeconomic, and commuting characteristics on mobility patterns. This process demonstrates the ability of the models to downscale from a larger spatial unit (taxi zone) to a smaller spatial unit (census tract).

4.5. Interpret the Blackbox
Although many of the machine learning non-linear models are often considered as "black box" models due to their complexity and nonlinear structure. This section aims to interpret our non-linear models by identifying the most influential independent variables with a perturbation-based sensitivity method. We seek to understand which demographic, socioeconomic, and commuting variables exhibit the strongest correlations with the number of taxi trips.

To achieve this, we adopt a perturbation-based sensitivity analysis. We systematically alter each independent variable slightly while holding all other independent variables constant and observe the resulting change in the model's output. For example, in the first iteration, we slightly alter the value of the origin's total population while keeping all other variables unchanged, resulting in a new predicted output $Y_1$. We then compute this change as:

$$\Delta Y_1 = Y_1 - Y, \quad \text{Eq. 5,}$$

where Y is the model's original output without perturbation. In the next iteration, we keep origin total population unchanged, but perturb another variable, such as destination total population, and calculate Y2 in the same manner. This process is repeated for each independent variable, producing a set of output changes

$$\Delta Y_i \; for \; i = 1,2,\dots,n,$$

where n is the number of independent variables.

Each $\Delta Y$ can be viewed as a pseudo-coefficient, similar to the coefficient in the linear regression. In this way, we can interpret non-linear models as linear regressions. Each $\Delta Y$ consists of two parts, the sign and the magnitude. A positive sign indicates positive correlation, meaning that an increase in the independent variable leads to an increase in predicted number of trips; while a negative sign indicates negative correlation, meaning that a decrease in the independent variable results in an increase in the number of trips. The magnitude reflects the sensitivity of the model output to changes in that specific independent variable.

Formally, for each independent variable $X_i$ we compute,

$$E_{X^{\text{ts}}} \left[ \frac{dY}{dX_i} \bigg|_{X=X^{\text{test}}} \right] \quad \text{Eq. 6.}$$

This allows us to quantify and rank the importance of all independent variables based on the magnitude of their associated $\Delta Y$, providing insights into the "black box" of the non-linear models.

To ensure consistency and comparability across the models used in this study, we apply the same perturbation-based sensitivity analyses to all four models: LR, RF, SVM, and FNN. For LR, the results of the sensitivity analysis are consistent with the coefficient directly obtained from the model. For the other three non-linear models, this method enables the extraction of interpretable correlations between independent variables and model predictions of the models. By systematically perturbing each independent variable and measuring the corresponding change in the predicted output, we effectively derive a comparable set of "pseudo-coefficients" for each model. This makes it possible to compare the importance of variables across different modeling approaches. This unified analysis allows us to evaluate and rank the influence of demographic, socioeconomic, and commuting-related variables revealed by each model, as well as compare the similarities and differences identified by different models.

5. **Results and Discussions**

In this section, we evaluate and compare the performance of four models in downscaling OD flows. Each model is assessed based on its predictive accuracy, as well as the correlations between demographic, socioeconomic, and commuting-related variables with the number of trips between each OD pair. Table 2 shows the normalized MSE and prediction RMSE when we apply the trained optimal models on the testing datasets. The FNN model achieves the best performance on the testing dataset, with the lowest normalized MSE at 0.118. The prediction RMSE of the FNN model is about 5672.849 trips. This is followed by the RF model, with a normalized MSE at 0.251 and prediction RMSE at 8291.179 trips. The SVM model shows larger normalized MSE at 0.365 and prediction RMSE at 9989.046. However, compared to the three non-linear models, the LR model shows high normalized MSE at 0.890 and prediction RMSE at 15602.30 trips, indicating the limitations of linear models.

Table 2. Normalized MSE and the trip prediction RMSE for testing dataset.

| Method | Z-score Zone MSE | Zone RMSE (trips) |
|---|---|---|
| Linear Regression | 0.890 | 15602.302 |
| Random Forest | 0.251 | 8291.179 |
| Support Vector Machine | 0.365 | 9989.046 |
| Feedforward Neural Network | 0.118 | 5672.849 |

Table 3 shows the normalized MSE and the prediction RMSE when we apply the trained optimal models for spatial downscaling. Compared to the three non-linear models, the LR shows extremely high errors in both normalized MSE and the prediction RMSE, meaning that the LR model does not work well on spatial downscaling. Comparing the three non-linear models, the SVM model performs best, with the smallest normalized MSE at 0.652 and the smallest prediction RMSE at 1043.07. The RF model shows higher errors than the SVM model, with the normalized MSE at 0.786 and prediction RMSE at 1144.517. However, FNN has the highest errors among the three non-linear models, with a normalized MSE at 0.844 and prediction RMSE at 1186.668. This result shows that the FNN model may overfit the training dataset, which results in a worse performance when applied to a new dataset.

Table 3. Normalized MSE and the trip prediction RMSE for spatial downscaling.

| Method | Z-score tract MSE | Tract RMSE (trips) |
|---|---|---|

| Linear Regression | 1.18E+23 | 4.43E+14 |
|---|---|---|
| Random Forest | 0.786 | 1144.517 |
| Support Vector Machine | 0.652 | 1043.067 |
| Feedforward Neural Network | 0.844 | 1186.668 |

In addition, error maps for RF, SVM, and FNN are provided to estimate the spatial distributions of predictive errors. These maps contain the unified color classifications, which means that the same color in different maps represents the same numeric range. This provides direct visual comparison among different models.

5.1. Regression

The linear regression (LR) model is a simple and interpretable approach that is commonly used to identify correlations between independent and dependent variables. While it offers direct interpretation of variable effects through coefficients, its limited ability fails to capture the complex correlations and nonlinear interactions in the OD flows.

In terms of predictive performance, linear regression performs significantly worse than other three nonlinear models. With a Z-score normalized MSE of $1.18 * 10^{23}$, the linear regression model's prediction deviates from the true values by $1.18 * 10^{23}$ standard deviations. Furthermore, the RMSE of $4.43 * 10^{14}$ trips indicates that the predicted trip counts differ from the observed values by over $4.43 * 10^{14}$ trips per OD pair on average. This high RMSE is likely due to the model's inability to account for interactions among variables and nonlinearities. By assuming only linear relationships exist, the LR model generates absurd errors, meaning that the linear model is effectively totally collapsing.

Since the LR model directly generates coefficients for each independent variable, it enables us to validate the perturbation-based sensitive analysis. In the case of the LR model, the sensitivity analysis produces results consistent with LR coefficients, reinforcing the validity of our perturbation approach as a method for identifying the associations between independent and dependent variables.

Among the independent variables, the most influential positive predictor is the number of commuters at the destination (LR coefficient = 0.269). This aligns with expectations, as tracts with a high density of employment opportunities, such as those in Downtown and Midtown Manhattan, attract more inbound trips. In addition, these locations are also associated with higher income levels, which may influence the choice of taxis as a preferred model of daily travel. The percentage of motorcycle commuters and the percentage of Hispanic population at the destinations also show positive associations with OD trip volumes. Additionally, the percentage of commuters with travel time longer than 90 minutes at the destination tract also displays a strong positive correlation, possibly reflecting that those tracts attract long-distance commuters using taxi. Notably, the most influential positive variables are all associated with the destination and most of those variables are related to commuting behaviors, suggesting that, according to the LR model, characteristics of the destination show stronger influence on trip volumes than those or the origins. This pattern also implies that many of the observed trips are likely work-related, reflecting flows toward employment centers during commuting times.

On the other hand, several variables on the origin side exhibit extreme negative coefficients, raising concern about the model's validity. For example, the total population at the origin tract has a correlation at $-2.38 * 10^{12}$, an implausibly large value that is clearly unrealistic. Similarly, variables such as the percentage of walk commuters, drive-alone commuters, and residents who have not moved in the last 12 months also show extremely large negative coefficients. These results suggest severe multicollinearity, where strong internal correlations among independent variables distort the model's predictions. For example, the percentage of commuters who drive alone and the percentage of walk commuters are likely to have strong negative correlations. This multicollinearity not only inflates the variance of coefficient estimates, but also undermines the model's ability to provide meaningful interpretations.

While the LR model indicates that the commuting-related variables have the most significant influences on the taxi OD flows, the extreme coefficient values and the substantial prediction errors limit the model's utility. Although variables such as the number of destination commuters and commuting behaviors are likely relevant, their effects cannot be reliably evaluated using the LR model in this context. Given the unreliability and the interpretative limitations of this model, we do not provide maps for the spatial errors, as we do not consider it to produce a meaningful or reliable downscaling result. These findings highlight the need for more advanced modeling approaches that can better handle the correlations among independent variables and nonlinear relationships.

5.2. Random forest

The random forest (RF) model offers a robust, non-parametric approach to modeling OD taxi flows, capable of capturing complex interactions and nonlinear relationships among demographic, socioeconomic, commuting-related features and the number of OD flows. In this study, the final RF model is constructed using two key hyperparameters: n_estimators = 10 and max_features = log2. The n_estimators parameter determines the number of individual decision trees in the ensemble. In this case, the RF model aggregates predictions from 10 trees, each trained on a different bootstrap sample of the training data. The max_features specifies that, at each decision split within a tree, the algorithm will randomly select a subset of features equal to the base-2 logarithm of the total number of features. Our hyperparameter tuning process indicates that these two settings result in less correlated trees, which promotes model diversity and helps to prevent overfitting.

After we train the model using 80% of the total OD flows at the taxi-zone level, we test the RF model on the remaining 20% of taxi-zone level flows. The RF model achieves a Z-score normalized MSE of 0.251 and an RMSE of 8,291 trips. These results indicate a substantial improvement from the regression model. When applying the trained RF model to the downscaling task, it achieves a Z-score normalized MSE of 0.785 and an RMSE of 1,144 trips. The increase in normalized MSE is expected, as it is applied to a new dataset. While these errors are higher than in the testing phase, the performance of the RF model remains acceptable given the complexity in the nature of human mobility.

A perturbation-based sensitivity analysis is applied to interpret the RF model and to identify which independent variables most strongly influence on the OD trip volumes. This method generates a set of pseudo-coefficients that reflect the model's reliance on each variable. The results show that several origin-side variables have the strongest positive correlation with the taxi flow volumes, including the percentage of carpool commuters (0.097) and percentage of public transit commuters (0.092). These two variables suggest that individuals who do not regularly use personal vehicles may rely more on taxis for daily travel. While the no-car percentage at the origin is positively correlated, its influence is weaker (0.018). In addition, the percentage of empty housing units (0.072) and the percentage of renter occupied units (0.056) also show strong positive correlations with flow volumes. These patterns may reflect residential instability or higher density housing environments, both of which increase reliance on taxis. On the other hand, several variables exhibit strong negative correlations with OD flow volumes. The most negatively correlated variable is the percentage of commuters using bicycles, suggesting that a bicycle-friendly environment may reduce reliance on taxis. Other important negatively correlated variables include the percentage of below-poverty population at the origin and the unemployment rate at the destination. These patterns likely reflect the economic constraints in low-income areas, where residents may avoid higher-cost transportation options in favor of public transit.

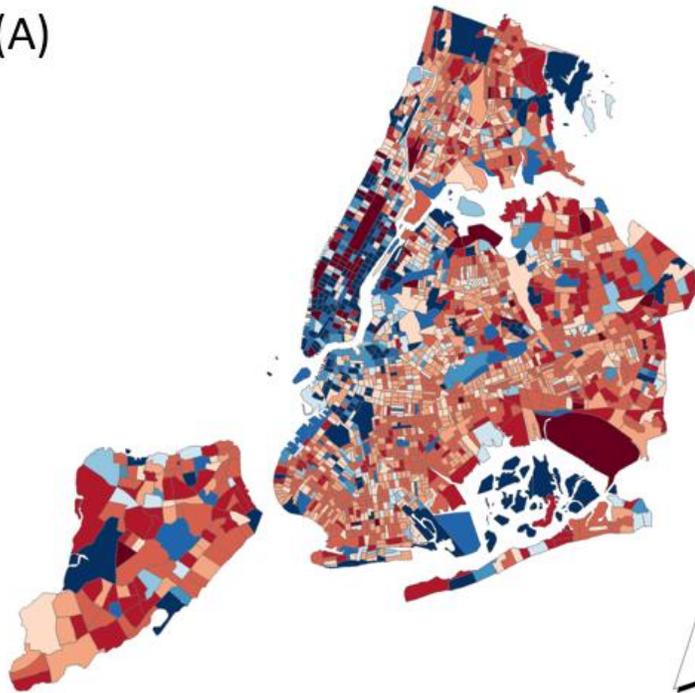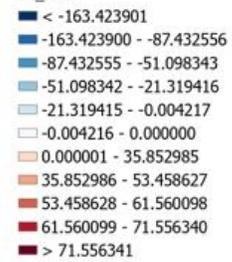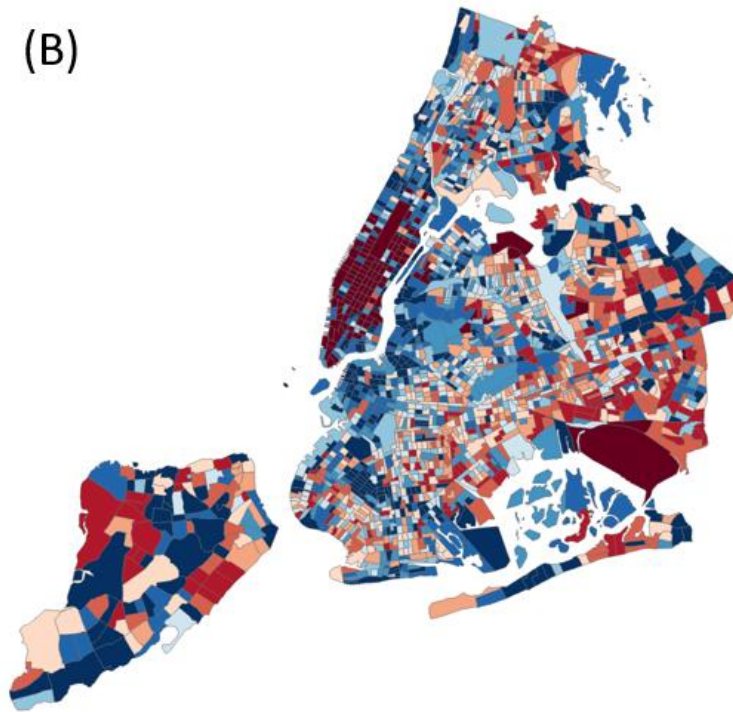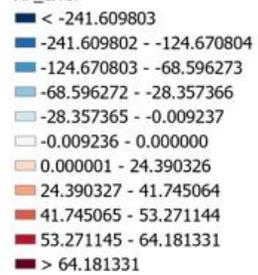

Figure 3. Error map for Random Forest model. (A) shows the downscaling error distribution at tract-level when each tract is considered as an origin. (B) shows the downscaling error distribution at the tract-level when each tract is considered as a destination.

To better understand how errors are distributed across space, we generate two spatial error maps: one showing prediction errors where each tract serves as an origin (Figure 3A), and the other where each tract serves as a destination (Figure 3B). To ensure comparability across all models, a unified classification scheme was employed, meaning that the same color in different maps represents the same numeric range. The origin error map reveals that tracts with higher origin-side errors are often located in the residential areas such as Staten Island, and parts of Queens and Brooklyn. These areas are characterized by greater automobile dependency, lower population density, and heterogeneous population composition. In general, the RF model tends to underestimate outbound flows, as indicated by the prevalence of red-shaded tracts. The destination error map presents a somewhat different pattern. Many tracts in Manhattan appear in dark red, suggesting that the model significantly underestimated inbound taxi trips into those areas. Conversely, tracts in Upper Manhattan and Western Brooklyn are mostly appearing blue-shaded, indicating overestimation of trips to those destinations. Notably, in both maps, Central Park, JFK airport, and LaGuardia Airport tracts appear in dark red, highlighting the model's underestimation of flows at these major non-residential areas. This may result from missing ACS data, as they do not have permanent residents.

In summary, the RF model demonstrates strong predictive capability, with solid performance on both the testing dataset with taxi zones, and the downscaled tract-level application. It successfully captures key features associated with OD flow volumes through perturbation-based sensitivity analysis. However, spatial error maps indicate that it still faces challenges, particularly in predicting flows at important non-residential areas.

5.3. SVM

The Support Vector Machine (SVM) model offers a robust nonlinear approach capable of capturing complex interactions in OD flow data without requiring strong assumptions about the relationships between variables. In this study, the final SVM model is developed using two hyperparameters: C = 10 and epsilon = 0.1. The regularization parameter C controls the trade-off between minimizing training error and maintaining a smooth model to prevent overfitting. The epsilon parameter defines a margin of tolerance within which no penalty is assigned to prediction errors. Among all the combinations tested during the hyperparameter tuning stage, this combination yields the lowest training error.

Trained on 80% of the taxi-zone-level dataset, the SVM achieved a Z-score normalized MSE of 0.365 and an RMSE of approximately 9,989 trips on the testing dataset. These results indicate moderate prediction accuracy during the model training and testing with the taxi-zone-level dataset. Although these testing errors are larger than the RF and NN model, when applied to the tract-level data for downscaling, the SVM model achieves the smallest prediction errors, with a Z-score normalized MSE of 0.652 and an RMSE of 1,043 trips. Compared to the testing dataset, this increase in MSE is expected due to the change of spatial resolution. Nevertheless, the model achieves the best downscaling performance among the three nonlinear models, highlighting its ability to capture complex spatial interactions in OD flows.

To further interpret the SVM model, a perturbation-based sensitivity analysis is conducted to derive a set of pseudo-coefficients. Among all the demographic, socioeconomic, and commuting-related variables, the total number of commuters at the destination shows the strongest positive correlation (0.069), emphasizing the role of employment centers in attracting taxi flows. Other variables with strong positive associations include the percentage of foreign-born population at the destination (0.057), the Hispanic population percentage at the origin (0.050) and the percentage of residents who did not move in the last year (0.050). On the other hand, the percentage of Hispanic population at the destination (-0.053) shows the strongest negative correlation. Taken together, these pseudo-coefficients indicate that the SVM model is particularly sensitive to nonlinear interactions between population structures.

Figure 4A and 4B. present the spatial distribution of downscaling prediction errors, with Figure 4A showing errors where each tract is the origin and Figure 4B where each tract is the destination. In both maps, the predominance of blue-shaded tracts indicates that the SVM model tends to overestimate the number of taxi trips in both directions. In Figure 4A, a cluster of red-shaded tracts in central Brooklyn reveals that the model underestimates the outbound trips in predominantly residential neighborhoods characterized by a racially and ethnically diverse population living in historical row houses. In contrast, tracts in eastern Queens and southern Staten Island, largely composed of low-density, single-family homes, are mostly shown in blue, indicating overestimation by the SVM model. These errors may reflect the model's limited ability to capture heterogeneous lifestyles that vary with population density, neighborhood structure, and local cultures. For errors at the destinations, the SVM model significantly underestimated inbound flows into downtown and midtown Manhattan, areas that serve as financial, commercial and tourism centers. These deep red zones raise concerns about the mode's challenge in capturing flows into non-residential but high-activity areas. This limitation is likely due to the nature of the ACS database, where demographic, socioeconomic, and commuting variables are tied to residential locations. This ignores key indicators for locations that attract non-resident visitors. Similar underestimations are observed for Central Park, JFK airport, and LaGuardia airport, again, reflecting the absence of residential data for these major travel destinations.

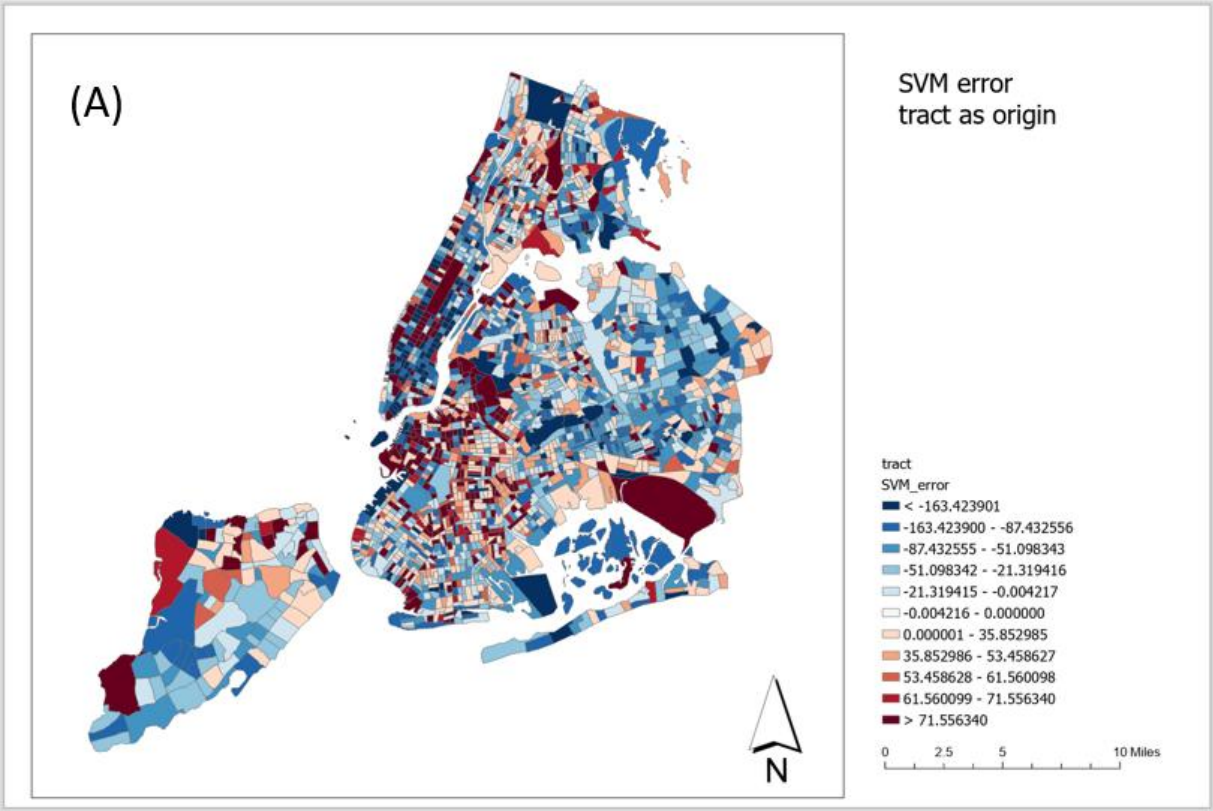

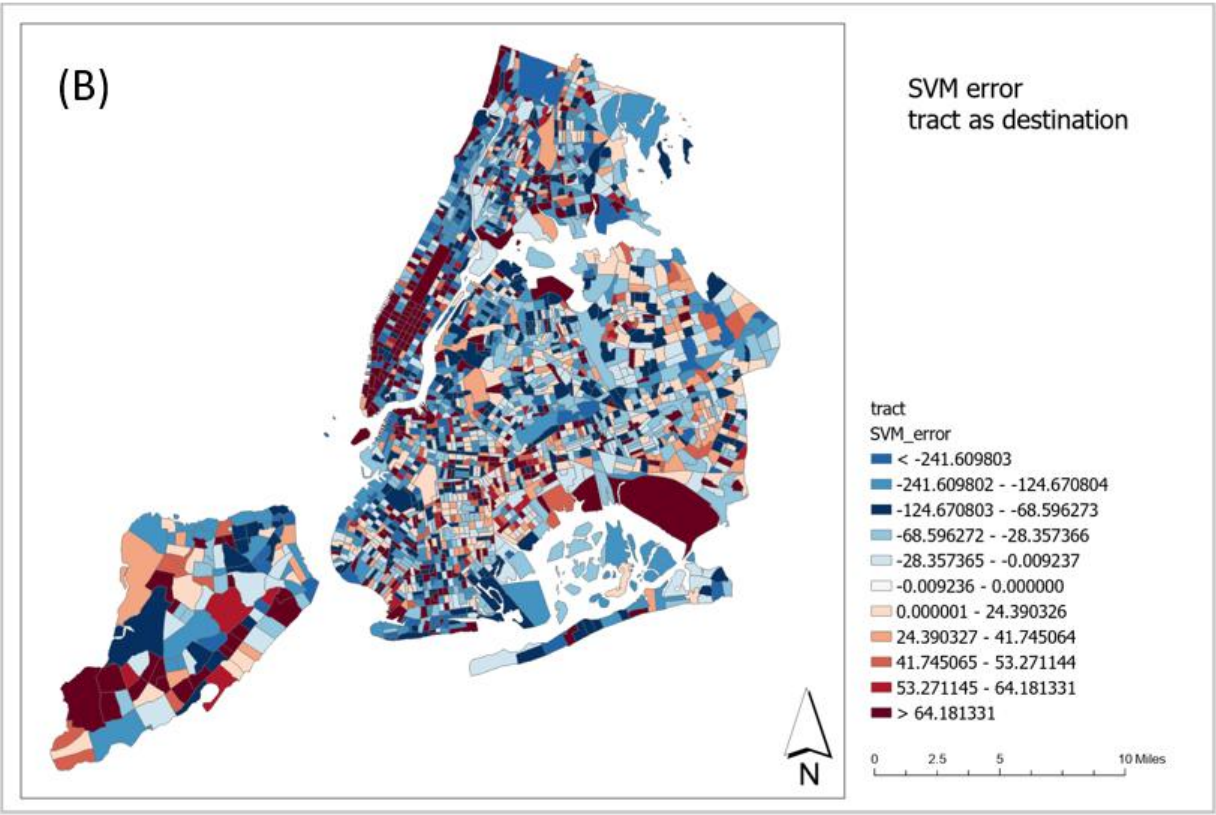

Figure 4. Error map for Support Vector Machine model. (A) shows the downscaling error distribution at tract-level when each tract is considered as an origin. (B) shows the downscaling error distribution at the tract-level when each tract is considered as a destination.

In summary, the SVM model demonstrates strong ability for downscaling OD flows to finer spatial units, with the best performance among the three nonlinear models. The perturbation-based sensitivity analysis indicates that it is particularly sensitive to demographic structure, but the prediction error maps show its limitation in capturing more complex OD flows across diverse urban environments.

5.4. Neural Network

The neural network (NN) model in this study uses a feedforward neural network, offering a flexible and powerful framework for capturing high-dimensional nonlinear interactions within complex datasets. After hyperparameter tuning, the final NN model consists of two hidden layers (height), each with 150 neurons (width).

During the training, the NN is trained using 80% of the OD taxi-zone-level flows. It achieved a z-score normalized MSE of 0.118 and an RMSE of 5672 trips. Among the three nonlinear models, the NN demonstrates the lowest MSE and RMSE on the testing dataset. However, when this NN model is applied to downscaling at the tract-level dataset, the Z-score normalized MSE is 0.844 and the RMSE is 1,186 trips, which is higher than both the RF and the SVM models. This decline in accuracy suggests that although the NN effectively captures patterns in the training dataset, it is less robust when generalizing to a different spatial resolution.

Similar to the other nonlinear models, a perturbation-based sensitivity analysis is conducted to understand the associations between independent variables and the dependent variable. Among all the features, the total number of commuters at the destination (0.055) has the strongest positive correlation, emphasizing the importance of work-related travel demand in influencing taxi flows. Other notable variables with positive correlations are the median income at the origin (0.046) and the male percentage at the destination (0.043), reflecting the influence of financial affordability and gender-based travel behaviors. On the other hand, the variable has the strongest negative correlation is the percentage of population that did not move in the past year at the destination (-0.041), followed by the percentage of commuters with more than 90-minute commuting time at the origin (-0.039) and the percentage of the work-from-home population at the origin (-0.029). These findings suggest that residential stability, long-distance commuting, and remote work are all associated with reduced taxi usage. In addition, the percentage of population without private vehicles at the destination (-0.028) also exhibits a strong negative correlation, indicating that residents in these areas are less likely to use taxi services because they already own private vehicles for their daily travel needs.

Figure 5A presents the spatial distribution of prediction errors where each tract is an origin, while Figure 5B shows errors where each tract is a destination. Visual inspection reveals that both maps are dominated by red-shaded areas, indicating that the NN model tends to underestimate the number of taxi flows. Such underestimation is particularly severe in Figure 5A, where most tracts in Staten Island and Queens are shown in deep red, highlighting a systematic underestimation of outbound trips from these residential areas. While underestimation persists for the inbound trips (Figure 5B), the magnitude of error appears somewhat lower. Notably, in both maps, the NN model overestimates the flows at Manhattan, as shown in both figures that many tracts in uptown and downtown Manhattan are in deep blue. These overestimations may reflect the NN model's difficulty in capturing the more complex and dynamic travel behaviors. Similar to other models, the NN also significantly underestimated flows at central park, JFK airport and LaGuardia airport, likely due to the lack of ACS data at these non-residential areas.

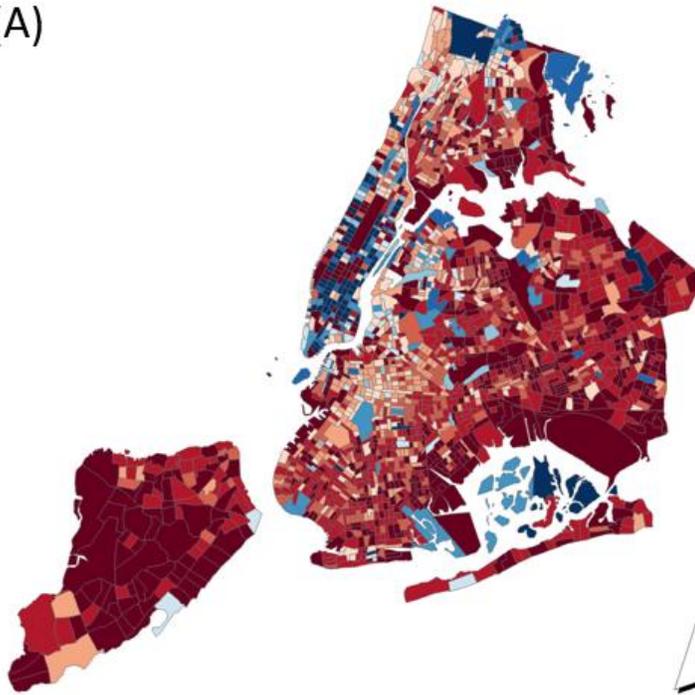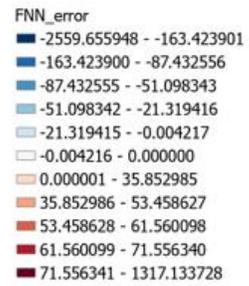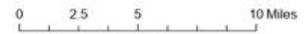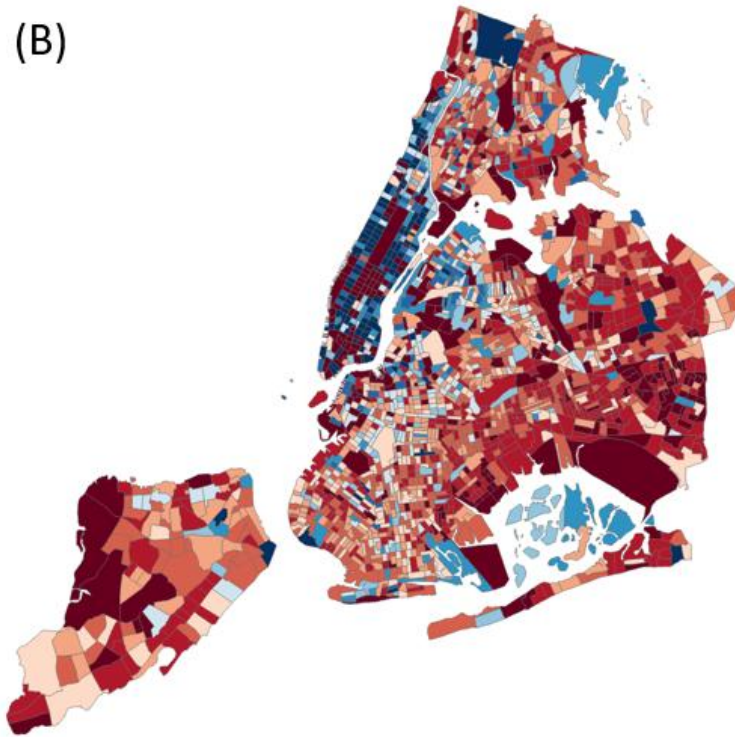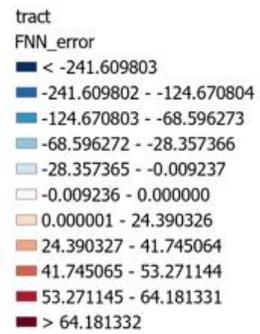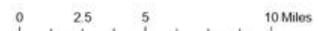

Figure 5. Error map for Neural Network model. (A) shows the downscaling error distribution at tract-level when each tract is considered as an origin. (B) shows the downscaling error distribution at the tract-level when each tract is considered as a destination.

In summary, the NN demonstrates high prediction accuracy during the testing phase with taxi-zone-level data, where patterns are more similar to the training dataset. However, when applied to the tract-level for downscaling, the model's performance declines. It systematically underestimates flows in residential boroughs, such as Staten Island and Queens, and overestimates in some parts of Manhattan, suggesting limitations in capturing the spatial and behavior complex of human movements at a finer spatial scale.

## 6. Discussion and Conclusion

This study presents a comprehensive framework for modeling and downscaling origin-destination (OD) taxi flow patterns in New York City by integrating trip records with demographic, socioeconomic, and commuting-related characteristics. We train four models—linear regression (LR), random forest (RF), support vector machine (SVM), and neural network (NN)—using OD flows at the taxi zone level, which represents the coarser spatial unit. These models are then applied to the census tract level, a finer spatial unit, to evaluate their downscaling performance. The LR model fails to capture the complex interactions among the variables and produces large errors, highlighting its limitations for this task. Among the three nonlinear models, the NN achieves the lowest error on the testing dataset, demonstrating strong learning capacity. However, it performs the worst during the downscaling stage, suggesting limited generalizability across spatial scales. In contrast, the SVM model achieves the best performance in the downscaling task, indicating robust spatial generalization ability. These results highlight the trade-offs between models' fitting accuracy and transferability, emphasizing the importance of model selection based on specific research goals.

To interpret the nonlinear models, we apply a perturbation-based sensitivity analysis that identifies how each independent variable influences the number of predicted taxi flows. This technique perturbs each variable slightly, one at a time, and measures the impact on the model's predictions, thereby estimating the relative importance of each variable. Across the three models, commuting behaviors consistently show strong correlations with taxi flow volumes. However, each model responded differently to other features: the RF model is more sensitive to the housing and economic conditions, the SVM model highlighted population composition variables, and the FNN model captures a broader mix of gender, income, commuting patterns, and housing stability. Spatial error maps at the tract level further reveal model-specific limitations. The NN model tends to systematically underestimate the flow volumes, particularly in residential areas, such as Staten Island and Queens. The SVM model, on the other hand, overestimates flow volumes in a large number of tracts. All three models exhibit large errors in Manhattan. This is partly due to the high volume of taxi trips there and partly because many of these trips involve non-residents traveling for work or tourism, which are not captured by ACS-based residential variables. Likewise, significant underestimations are found in central park, JFK airport, and

LaGuardia airport, likely due to a mismatch between ACS data and frequent movements in these non-residential locations.

Despite its contributions, this study faces several limitations. Most notably, the ACS datasets only provide residential characteristics, limiting the model's ability to capture trips with non-residential purposes. In areas like downtown and midtown Manhattan, a substantial portion of taxi flows can be for work-related (e.g., between office buildings), or tourism-driven (e.g., visits to landmarks), which cannot be fully explained by residential characteristics alone. Future research should incorporate additional data sources, such as land use, Points-of-Interests, and workplace-related statistics to enhance model accuracy. Second, this study only uses taxi trip records. In order to better capture broader human movements across the city, future studies should consider data-fusing with different types of human mobility data types, such as cellphone-based mobility data. Third, temporal dynamics in mobility are not addressed in this study. For example, the 2017 statewide legalization of ridesharing services in New York State significantly changed the mobility patterns in NYC. Similarly, events such as the COVID-19 pandemic also dramatically reshaped movement patterns. Models trained on 2011 data cannot account for such changes. Future studies should explore how to adapt models to transportation-related policy changes. Lastly, because the models in this study are trained specifically on NYC data, their applicability to other cities remains uncertain. Developing more transferable models for broader geographic use is an important direction for future research.

In conclusion, this study makes three main contributions to the understanding of human mobility patterns in urban environments. First, it introduces an interpretable GeoAI machine learning framework for downscaling origin-destination (OD) flows from a coarse spatial unit (taxi zones) to a finer scale unit (census tracts), addressing a common challenge in spatial movement analysis. Second, it enhances model interpretability by applying the perturbation-based sensitivity analysis to interpret how each independent variable influences the dependent variable. This method partially mitigates concerns of viewing GeoAI models as black-boxes. Third, it integrates detailed demographic, socioeconomic, and commuting-related variables with mobility records, enabling interpretable, fine-scale insights into the factors that influence taxi usage. The methodology presented in this study provides both analytical advancement and practical utilities. For researchers, the downscaling method opens up new opportunities to explore urban dynamics at a finer spatial resolution than what is directly available from the raw datasets. For urban developers and transportation planners, the findings provide a more localized understanding of travel behavior, supporting data-driven planning and resources allocation.

**Availability of data and material:**
All the datasets used in this study are open access data and are listed in reference [78] and [79].

**Competing Interests:**
The authors declare no competing interests.

**Funding:**
This study is funded by University of Hawaiʻi at Mānoa College of Social Science MA-3802347.

**Authors' contributions:**
Y.J., A.A.P. and Q.L. designed this study. All authors participated in the experiment and results analysis. Y.J. and A.A.P. created all the figures and tables. Y.J., A.A.P. and T.D. prepared the manuscript. All authors reviewed the manuscript.

**Acknowledgement:**
Y.J. and A.A.P. would like to thank the endless roads in Texas for their unexpected role in inspiring this research idea.